\documentclass[11pt]{article}
\usepackage{eamt23}
\usepackage{times}
\usepackage{url}
\usepackage{latexsym}
\usepackage[small,bf]{caption} 
\usepackage{amsmath}
\usepackage{hyperref}
\usepackage{tablefootnote}

\usepackage{float}

\usepackage[utf8]{inputenc}

\usepackage{graphicx}
\usepackage{dsfont}
\usepackage{booktabs,arydshln}

\usepackage{microtype}
\usepackage{amssymb}
\usepackage{multirow}

\usepackage{balance}
\usepackage{booktabs}
\usepackage[normalem]{ulem}
\usepackage{enumitem}
\usepackage{subcaption}
\usepackage{inconsolata}
\usepackage{changes}

 \usepackage{color}
 \usepackage{xcolor}

\newcommand{\tagok}{\textsc{ok }}
\newcommand{\tagbad}{\textsc{bad } }

\newcommand{\comet}{\textsc{Comet }}

\newcommand{\chrf}{\textsc{chrF }}

\title{BLEU Meets \textsc{Comet}: Combining Lexical and Neural Metrics \\ Towards Robust Machine Translation Evaluation}

\author{
Taisiya Glushkova$^{1, 3}$ 
Chrysoula Zerva$^{1, 3}$ 
André F. T. Martins$^{1, 2, 3}$ 
\\
$^1$Instituto de Telecomunicações \quad $^2$Unbabel \quad  \\
$^3$Instituto Superior Técnico \& LUMLIS (Lisbon ELLIS Unit) \\
{\small \texttt{\{taisiya.glushkova, chrysoula.zerva, andre.t.martins\}@tecnico.ulisboa.pt}}\\
}

\date{}

\begin{document}
\maketitle
\begin{abstract}

Although neural-based machine translation evaluation metrics, such as {\sc Comet} or {\sc BLEURT}, have achieved strong correlations with human judgements, they are sometimes unreliable in detecting certain phenomena that can be  considered as critical errors, such as deviations in entities and numbers. 
In contrast, traditional evaluation metrics, such as {\sc BLEU} or {\sc chrF}, which measure lexical or character overlap between translation hypotheses and human references, have lower correlations with human judgements but are sensitive to such deviations. 
In this paper, we investigate several ways of combining the two approaches in order to increase robustness of state-of-the-art evaluation methods to translations with critical errors. We show that by using additional information during training, such as sentence-level features and word-level tags, the trained metrics improve their capability to penalize translations with specific troublesome phenomena, which leads to gains in correlation with human judgments and on recent challenge sets on several language pairs.\footnote{Our code and data are available at: \url{https://github.com/deep-spin/robust_MT_evaluation}}

\end{abstract}

\section{Introduction}
\label{sec:intro}

Trainable machine translation (MT) evaluation models, such as \textsc{Comet} \cite{rei-etal-2020-comet} and \textsc{BLEURT} \cite{sellam-etal-2020-bleurt}, generally achieve higher correlations with human judgments, thanks to leveraging pretrained language models. However, they often fail at detecting certain types of errors and deviations from the source, for example related to translations of numbers and entities \cite{amrhein2022identifying}. 
As a result, their quality predictions are sometimes hard to interpret and not always trustworthy. 
In contrast, traditional lexical-based metrics, such as \textsc{BLEU} \cite{papineni-etal-2002-bleu} or \textsc{chrF} \cite{popovic-2015-chrf}---despite their many limitations---are considerably more sensitive to these errors, due to their nature, and are also more interpretable, since the scores can be traced back to the character or $n$-gram overlap.

This paper investigates and compares methods that combine the strengths of neural-based and lexical approaches, both at the sentence level and at the word level. 
This is motivated by the findings of 
previous works, which demonstrate in detail that the {\sc Comet} MT evaluation metric struggles to handle errors like deviation in numbers, wrong named entities in generated translations, deletions that exclude important content from the source sentence, insertions of extra words that are not present in the source sentences, and a few others \cite{amrhein2022identifying,alves-etal-2022-robust}. While data augmentation techniques alleviate the problem to some extent \cite{alves-etal-2022-robust}, the gains seem to be relatively modest. 
In this paper we investigate alternative methods that take advantage of lexical information and go beyond the use of various augmentation techniques and synthetic data. 

We focus on increasing robustness of MT evaluation systems to certain types of critical errors. We experiment with the reference-based {\sc Comet} metric, which has access to reference translations when producing quality scores. 
In order to make evaluation metrics more robust towards this type of errors, we consider and compare three different ways of incorporating information from lexical-based evaluation metrics into the neural-based {\sc Comet} metric:
\begin{itemize}
\setlength\itemsep{0em}
\item Simply ensembling the sentence-level metrics;
\item Using lexical-based sentence-level scores as additional features through a bottleneck layer in the {\sc Comet} model;
\item Enhancing the word embeddings computed by {\sc Comet} for the generated hypothesis with word-level tags. We generate these word-level tags using the Levenshtein (sub)word alignment between the hypothesis and the reference tokens.
\end{itemize}

We compare these three strategies with the recent approach of \cite{alves-etal-2022-robust}, which generates synthetic data with injected errors from a large language model, and retrains {\sc Comet} on training data that has been augmented with these examples. We assess both the correlation with human judgments and using the recently proposed DEMETR benchmark \cite{karpinska-etal-2022-demetr}.

\section{Related Work}

Recently several challenge sets  have been introduced, either within a scope of the WMT Metrics shared task \cite{freitag-etal-2022-results} or in general as a step towards implementing more reliable MT evaluation metrics: SMAUG \cite{alves-etal-2022-robust} explores sentence-level multilingual data augmentation; ACES \cite{amrhein-etal-2022-aces} is a translation accuracy challenge set that covers high number of different phenomena and language pairs, including a considerable number of low-resource ones; DEMETR \cite{karpinska-etal-2022-demetr} and HWTSC \cite{chen-etal-2022-exploring} aim at examining metrics ability to handle synonyms and to discern critical errors in translations; DFKI \cite{avramidis-macketanz-2022-linguistically} employs a linguistically motivated challenge set for two language directions (German $\leftrightarrow$ English).

Apart from purely focusing on improving robustness with augmentation of different phenomena, there are works that combine usage of synthetic data with other different methods. These methods use more fine-grained information---aiming at identifying both the position and the type of translation errors on given source-hypothesis sentence pairs \cite{bao2023towards}.  
As another source of useful information, word-level supervision can be considered, which has proven to be beneficial in tasks of quality estimation and MT evaluation \cite{rei-etal-2022-comet,rei-etal-2022-cometkiwi}.

There have been other attempts to add linguistic features to automatic MT evaluation metrics, e.g. incorporating information from a  multilingual knowledge graph into BERTScore. \cite{wu2022kg} proposed a metric that linearly combines the results of BERTScore and bilingual named entity matching for reference-free machine translation evaluation. \cite{abdi2019deep} use an extensive set of linguistic features at word- and sentence- level to aid sentiment classification. 
Additionally, glass-box features extracted from the MT model have been used successfully in the quality estimation task \cite{fomicheva2020unsupervised,zerva2021unbabel,wang2021beyond}. For the incorporation of different types of information to neural models early and late fusion is commonly used with benefits on multiple tasks and domains \cite{gadzicki2020early,fu2015fast,baltruvsaitis2018multimodal}. To the best of our knowledge there have not been any attempts to combine the representations of neural metrics with external features obtained by lexical-based metrics.

Moreover, there are similar concerns regarding robustness of evaluation models in non-MT related tasks \cite{chen2022menli}. In general, it is depicted that evaluation metrics perform rather well on standard evaluation benchmarks but are vulnerable and unstable to adversarial examples. The approaches investigated in our paper aim to address these limitations.

\section{Combination of Neural and Lexical Metrics}

In this section we describe the methods we investigated in order to infuse the \textsc{Comet} with information on lexical alignments between the MT hypothesis and the reference. 

\subsection{Metric ensembling}

A simple way to combine neural and lexical-based metrics is through an ensembling strategy.
To this end, we use a weighted ensemble of normalized \textsc{Bleu}, \textsc{chrF} and \textsc{Comet} scores. The weights for each metric are tuned on the same development set used for training the \comet models discussed in this work (MQM WMT 2021) and presented in Appendix \ref{sec:app_model_params}. For normalisation we compute the mean and standard deviation to standardize the development set for each metric and we use the same mean and standard deviation values to standardize the test-set scores.  

\subsection{Sentence-level lexical features}
A simple ensemble is limited because it does not let the neural-based model \emph{learn} the best way of including the information coming from the lexical metrics---for example, the degree of additional information brought by the lexical metrics might depend on the particular input.

Therefore, we experiment with a more sophisticated approach, where the lexical scores are  incorporated in the \textsc{Comet} architecture as additional features that are mapped to each instance in the data, allowing the system to learn how to best take advantage of these features. 
To this end, we adopt a late fusion approach, employing a bottleneck layer to combine the lexical and neural features. The use of a bottleneck layer for late fusion in deep neural architectures has been used successfully across tasks, especially for multimodal fusion or fusion of features with vast differences in dimensionality \cite{petridis2017end,guo2018feature,ding2022multimodal}.
In our implementation, the bottleneck layer is inserted between two feed-forward layers in the original \textsc{Comet} architecture (see Fig.~\ref{fig:sl-arch}), implemented in a similar manner as in \cite{moura-etal-2020-ist,zerva-etal-2022-disentangling} (see App. ~\ref{sec:app_model_params}).

\subsection{Word-level lexical features}

While the sentence-level features allow the model to account for lexical overlap, there is still no word-level information. Instead, we propose to leverage the inferred alignments between the MT hypothesis and the reference words. To that end we adopt the Translation Edit Rate (TER) alignment procedure that calculates the edit distance (cost) between the translation and reference sentence. This alignment, produced with the Levenshtein dynamic programming algorithm, identifies the minimal subset of MT words that would need to be changed (modified, inserted, or deleted) in order to reach the correct translation (reference) \cite{snover2009ter}. TER-based alignments have been widely used to evaluate translations with respect to post-edits (HTER) in automated post-editing as well as other generation tasks \cite{snover2009fluency,elliott2014comparing,gupta2018deep,bhattacharyya2022findings}. Recently, providing word-level supervision using binary quality tags inferred via Multidimensional Quality Metrics (MQM)
error annotations, proved to be beneficial for MT evaluation \cite{rei-etal-2022-comet}.

In this work, for simplicity, we opted for calculating the alignments not on a word but on a sub-word level, employing the same tokenization convention used by the \textsc{Comet} encoder.\footnote{We specifically used the \href{https://huggingface.co/docs/transformers/model_doc/xlm-roberta\#transformers.XLMRobertaTokenizer}{\tt XLMRobertaTokenizerFast} Huggingface implementation with truncation and default \texttt{max\_length}.}  This allows to associate a quality \tagok / \tagbad tag to each sub-word unit of the MT hypothesis input vector. 

\begin{figure*}[t]
\centering
\begin{minipage}[r]{1.0\columnwidth}
    \centering
    \includegraphics[width=\columnwidth]{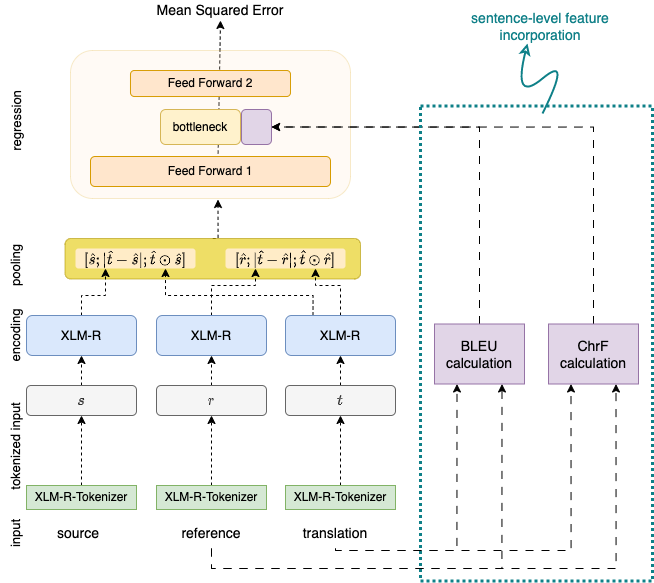}
    \caption{The architecture of the {\sc Comet} model with incorporated sentence-level lexical features.}
    \label{fig:sl-arch}
\end{minipage}
\hfill{}
\begin{minipage}[l]{1.0\columnwidth}
    \centering
    \includegraphics[width=\columnwidth]{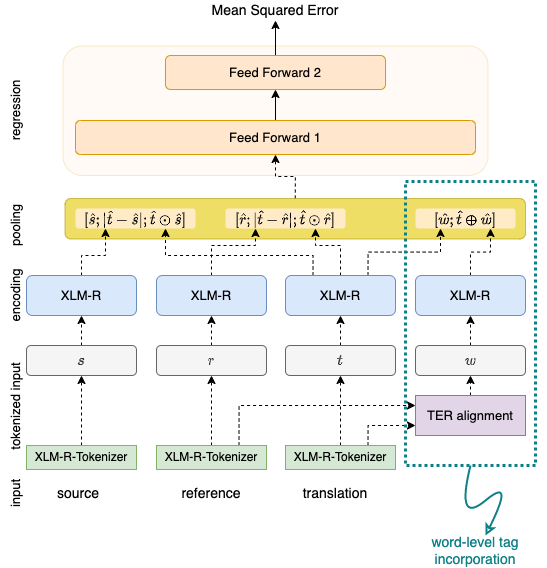}
    \caption{The architecture of the {\sc Comet} model with incorporated word-level lexical features.}
    \label{fig:wl-tag-arch}
\end{minipage}
\end{figure*}

We then incorporate these quality tags to the original input for each translation sample which consists of a triplet $\langle s, t, r\rangle$, where $s$ is a source text, $t$ is a machine translated text, and $r$ is a reference translation. 
To leverage the estimated quality tags in the \comet architecture, we encode the tags as a sequence of special tokens, $w$, and learn separate embeddings for the \tagok/ \tagbad  tokens. 
We can thus encode the quality tag sequence and obtain a word quality vector $\vec{w}$ and then compute the sum $\vec{\sigma} = \vec{t} \oplus \vec{w}$ for the sequence. We then extend the pooling layer of \textsc{Comet} by adding both the $\vec{w}$ and $\vec{\sigma}$ representations (see the architecture in Fig.~\ref{fig:wl-tag-arch}). 

\section{Experimental Design}
\label{sec:experiments}

The main focus of our experiments is to investigate how the robustness of the MT evaluation models can be improved and how the proposed settings compare to each other and to a data augmentation approach proposed by \cite{alves-etal-2022-robust}.
Our comparisons address the correlation with human judgments and recent robustness benchmarks on MT evaluation datasets (\S\ref{sec:data}).

We follow \cite{amrhein2022identifying} – we use \textsc{Comet} (v1.0) \cite{rei-etal-2020-comet} 
as the underlying architecture for our MT evaluation models and focus on making it more robust.

\paragraph{Human Judgements Data}
\label{sec:data}
We consider two types of human judgments: direct assessments (DA) \cite{graham-etal-2013-continuous} and multi-dimensional quality metric scores (MQM) \cite{lommel2014multidimensional}. 
For training, we use WMT 2017--2020 data from the Metrics shared task \cite{freitag-etal-2021-results} with direct assessment (DA) annotations (see App. ~\ref{sec:data_stats}). 
For development and test, we use the MQM annotations of the WMT 2021 and 2022 datasets, respectively~\footnote{We opted for DA annotations to train due to the limited availability of MQM data}.

\paragraph{Challenge Sets Data}
Furthermore, we evaluate our models using two challenge sets: DEMETR \cite{karpinska-etal-2022-demetr} and  ACES \cite{amrhein-etal-2022-aces}. 

\begin{itemize}
\setlength\itemsep{0em}
\item DEMETR is a diagnostic dataset with 31K English examples (translated from 10 source languages) created for evaluating the sensitivity of MT evaluation metrics to 35 different linguistic perturbations spanning semantic, syntactic, and morphological error categories. Each example in DEMETR consists of (1) a sentence in one of 10 source languages, (2) an English translation written by a human translator, (3) a machine translation produced by Google Translate, and (4) an automatically perturbed version of the Google Translate output which introduces exactly one mistake (semantic, syntactic, or typographical). 
\item ACES is a translation accuracy challenge set based on the MQM ontology. It consists of 36,476 examples covering 146 language pairs and representing 68 phenomena. 
This challenge set consists of synthetically generated adversarial examples, examples from repurposed contrastive MT test sets, and manually annotated examples. 
\end{itemize}

Both of these challenge sets allow measuring the  sensitivity of the proposed approaches to various phenomena and assess their overall robustness.

\paragraph{Augmentation}
We compare our methods against the multilingual data augmentation approach \textsc{SMAUG}\footnote{The code is available at \url{https://github.com/Unbabel/smaug}.} proposed by \cite{alves-etal-2022-robust}.
Specifically, we use transformations that focus on deviations in named entities and numbers since these are identified as the major weaknesses of \textsc{Comet} \cite{amrhein2022identifying}.

\paragraph{Models} 
In the experiments that follow, we use as baseline the vanilla \textsc{Comet} architecture trained on WMT2017--2020 (\textbf{\textsc{Comet}}). 
We compare this baseline against the model trained with augmented data and our proposed approaches:
\begin{itemize}
\setlength\itemsep{0em}
    \item \textbf{\textbf{\textsc{Comet}} + aug}: \textsc{Comet} model trained on a mixture of original and augmented WMT2017--2020 data, where the percentage of the augmented data is 40\%. We use the code provided by the authors of \textsc{SMAUG} and apply their choice of hyperparameters, including the optimal percentage of the augmented data. 

    \item \textbf{Ensemble}: The weighted mean of \textsc{Bleu}, \textsc{chrF} and \textsc{Comet} normalized scores, where the weights are optimized on the development set (MQM 2021) with regards to the Kendall's tau correlations.
    
    \item \textbf{\textbf{\textsc{Comet}} + SL-feat.}: The combination of \textsc{Comet} and scores obtained from other metrics, \textsc{BLEU} and \textsc{chrF}, that are used as sentence-level (SL) features in a late fusion manner.
    
    \item \textbf{\textbf{\textsc{Comet}} + WL-tags}: The combination of \textsc{Comet} and word-level \tagok/ \tagbad tags that correspond to the subwords of the translation hypothesis. 
    
\end{itemize}

\paragraph{Evaluation}
For evaluation and analysis we: 
\begin{enumerate}
    \item Compute standard correlation metrics on segment-level between predicted scores and human judgements: Pearson $r$, Spearman $\rho$ and Kendall's tau; 
    \item Use challenge sets, specifically DEMETR and ACES, to analyse the robustness of MT Evaluation systems to critical errors and specific perturbations. 
\end{enumerate}  

For the challenge sets, we measure the ability of the evaluation metric to rank the correct translations higher than the incorrect ones by computing the official Kendall's tau-like correlation as proposed in previous WMT Metrics shared tasks \cite{freitag-etal-2022-results,ma-etal-2019-results}:
\begin{equation}
    \tau = \frac{\text{Concordant} - \text{Discordant}}{\text{Concordant} + \text{Discordant}},
\end{equation}
where the ``{Concordant}'' is the number of times a metric assigns a higher score to the ``better'' hypothesis and ``Discordant'' is the number of times a metric assigns a higher score to the ``worse'' hypothesis. 

\section{Results and Discussion}

In this section, we show results for the aforementioned methods, specifically the correlations with MQM annotations from WMT 2022 Metrics shared task for 3 high-resource language pairs (English $\rightarrow$ German, English $\rightarrow$ Russian, Chinese $\rightarrow$ English) in four domains: Conversation, E-commerce, News and Social. 
In addition, we discuss evaluation results obtained on two challenge sets.
\begin{table*}[ht!]
\small
\centering
\begin{tabular}{clcccccccc}
\toprule
& & \multicolumn{1}{c}{\textsc{Bleu}} & \multicolumn{1}{c}{\textsc{chrf}} & \multicolumn{1}{c}{{\textsc{Comet}}} & \multicolumn{1}{c}{\textsc{ensemble}} & \multicolumn{1}{c}{\textsc{Comet}+aug} & \multicolumn{1}{c}{\textsc{Comet}+SL-feat.}  & \multicolumn{1}{c}{\textsc{Comet}+WL-tags}\\
\cmidrule{3-10}  
\multirow{4}{*}{\rotatebox{90}{\textsc{En-De}}} 
  & Conversation & 0.201 & 0.257 & 0.308 & 0.309 & 0.296 & 0.310  & \textbf{0.314}\\
  & E-commerce    & 0.179 & 0.212 & \textbf{0.326} & 0.318 & 0.311 & 0.322  & 0.322\\
  & News         & 0.167 & 0.202 & 0.361 & 0.356 & 0.330 & 0.355  & \textbf{0.369}\\
  & Social       & 0.130 & 0.168 & \textbf{0.297} & 0.292 & 0.277 & 0.294  & 0.293\\
  \midrule
\multirow{4}{*}{\rotatebox{90}{\textsc{En-Ru}}}  
  & Conversation & 0.140 & 0.175 & 0.305 & 0.304 & \textbf{0.328} & 0.298 & \textbf{0.328}\\
  & E-commerce    & 0.202 & 0.221 & 0.372 & 0.371 & 0.382 & 0.369 & \textbf{0.391}\\
  & News         & 0.125 & 0.164 & \textbf{0.373} & 0.367 & 0.366 & 0.384 & 0.370\\
  & Social       & 0.152 & 0.132 & 0.305 & 0.304 & 0.330 & 0.332 & \textbf{0.349}\\
  \midrule
  \multirow{4}{*}{\rotatebox{90}{\textsc{Zh-En}}}  
  & Conversation  & 0.125 & 0.160 & 0.283 & 0.282 & 0.295 & 0.283 & \textbf{0.298}\\
  & E-commerce     & 0.174 & 0.187 & 0.326 & 0.325 & 0.342 & 0.335 & \textbf{0.357}\\
  & News          & 0.046 & 0.042 & 0.270 & 0.261 & 0.291 & 0.276 & \textbf{0.292}\\
  & Social        & 0.162 & 0.190 & 0.319 & 0.316 & 0.313 & 0.315 & \textbf{0.330}\\
  \midrule 
\multirow{4}{*}{\rotatebox{90}{\textsc{ }}}  
  & AVG           & 0.150 & 0.176 & 0.321 & 0.317 & 0.322 & 0.323 & \textbf{0.334}$^\dagger$\\
\bottomrule
\end{tabular}
\addtolength{\tabcolsep}{-0.5pt}
\caption{Kendall’s tau correlation on high resource language pairs using the MQM annotations for Conversation, E-commerce, News and Social domains collected for the WMT 2022 Metrics Task. \textbf{Bold} numbers indicate the best result for each domain in each language pair. $\dagger$ in the averaged scores indicates statistically significant difference to the other metrics \tablefootnote{For the statistical significance over correlations $r$ we use Williams' test and Fisher $r-to-z'$ transform: $f(r) = \frac{1}{2}\ln{\frac{1+r}{1-r}}$ to calculate significance over the macro-averages, with $p<=0.01$.
}. }
\label{tab:mqm_kendall}
\end{table*}

\subsection{Correlation with Human Judgements}

Overall, by looking at Table~\ref{tab:mqm_kendall} we can see that the more sophisticated techniques of using additional information, whether it is lexical-based scores used as features, word-level tags based on token alignments or synthetically augmented data, outperform the simple weighted average (ensemble) approach. These findings are further supported when calculating performance for the Pearson $r$ and Spearman $\rho$ coefficients, shown in Tables \ref{tab:mqm_pearson} and \ref{tab:mqm_spearman} respectively in the Appendix \ref{sec:app_mqm}.

Across all proposed methods, we observe that \textbf{\textsc{Comet} + aug} and \textbf{\textsc{Comet} + SL-feat.} have relatively similar performance.  
In contrast, adding word-level tags (\textbf{\textsc{Comet} + WL-tags}) based on alignments between the translation hypothesis and the reference seems to give a considerable gain in results compared to the baseline \textbf{\textsc{Comet}} and the other approaches.

Another interesting observation is that the improvement in correlations can be especially noticed in \textsc{Zh-En} language pair across all domains for \textbf{\textsc{Comet} + WL-tags} model.
Overall, we found that adding the word-level quality supervision provides the most consistent benefits in performance. However, since our main motivation is to address robustness to specific errors, the correlations with MQM annotations serve primarily as a confirmation of the potential of the proposed methods; we provide a more detailed performance analysis over the multiple error types of different challenge sets in the next section. 

\subsection{Results on Challenge Sets}

\subsubsection{DEMETR}
For DEMETR we analyse results on two levels of granularity: (1) performance over the full challenge set, which is calculated via Kendall's tau and presented in Table~\ref{tab:demetr} which shows Kendall's tau-like correlations per language pair; and (2) performance depending on error severity, which is presented in and Table~\ref{tab:demetr2} and shows accuracy on detecting different types of DEMETR perturbations for lexical and neural-based metrics, bucketed by error severity (baseline, critical, major, and minor errors). 

\begin{table*}[ht!]
\small
\centering
\begin{tabular}{lcccccccc}
\toprule
& \multicolumn{1}{c}{\textsc{BLEU}} & \multicolumn{1}{c}{\textsc{chrF}} & \multicolumn{1}{c}{{\textsc{Comet}}} & \multicolumn{1}{c}{\textsc{Ensemble}} & \multicolumn{1}{c}{\textsc{Comet}+aug} & \multicolumn{1}{c}{\textsc{Comet}+SL-feat.} & \multicolumn{1}{c}{\textsc{Comet}+WL-tags} \\
\cmidrule{2-9}  
  \textsc{Zh-En}  & 0.505 & 0.684 & 0.818 & 0.855 & 0.817 & 0.866 & \textbf{0.872 }\\ 
  \textsc{De-En}  & 0.655 & 0.802 & 0.909 & 0.926 & 0.917 & 0.942 & \textbf{0.957} \\ 
  \textsc{Hi-En}  & 0.616 & 0.768 & 0.900 & 0.92 & 0.925 & 0.929 & \textbf{0.945} \\ 
  \textsc{Ja-En}  & 0.521 & 0.722 & 0.850 & 0.883 & 0.83 & \textbf{0.907} & 0.891 \\ 
  \textsc{Ps-En}  & 0.533 & 0.703 & 0.818 & 0.88 & 0.775 & 0.863 & \textbf{0.877} \\ 
  \textsc{Ru-En}  & 0.552 & 0.724 & 0.898 & 0.91 & 0.894 & \textbf{0.950} & 0.949 \\ 
  \textsc{Cz-En}  & 0.541 & 0.755 & 0.875 & 0.917 & 0.863 & 0.87 & \textbf{0.920} \\ 
  \textsc{Fr-En}  & 0.664 & 0.794 & 0.892 & 0.915 & 0.926 & 0.945 & \textbf{0.951} \\ 
  \textsc{Es-En}  & 0.516 & 0.704 & 0.877 & 0.899 & 0.877 & 0.91 & \textbf{0.934} \\  
  \textsc{It-En}  & 0.601 & 0.774 & 0.912 & 0.924 & 0.906 & 0.936 & \textbf{0.945} \\ 
  \midrule 
  AVG  & 0.57 & 0.743 & 0.875 & 0.903 & 0.873 & 0.912 & \textbf{0.924}$^\dagger$ \\ 
\bottomrule
\end{tabular}
\addtolength{\tabcolsep}{-0.5pt}
\caption{Kendall’s tau-like correlation per language pair on DEMETR challenge set. \textbf{Bold} values indicate the best performance per language pair. $\dagger$ in the averaged scores indicates statistically significant difference to the other metrics.}
\label{tab:demetr}
\end{table*}

We can observe that both the sentence- and word-level features outperform data augmentation methods, with the word-level method being the best on average and for the majority of language pairs. These findings indicate that the subword quality tags enable the model to attend more to the perturbations of the high quality data, hence better  distinguishing the bad from the good translations of the same source.

\begin{table}[ht!]
\small
\centering
\addtolength{\tabcolsep}{-0.5pt}
\resizebox{7.7cm}{!}{
\begin{tabular}{lccccc}
\toprule
Metric & Base & Crit. & Maj. & Min. & All \\  
\midrule
  \multicolumn{6}{c}{\textit{lexical-based metrics}} \\
  \textsc{BLEU} & \textbf{100.0} & 79.33 & 83.76 & 72.6 & 78.52 \\
  \textsc{chrF} & \textbf{100.0} & 90.79 & 90.85 & 80.83 & 87.16 \\
  \midrule
  \multicolumn{6}{c}{\textit{neural-based metrics}} \\
  \textsc{Ensemble}   & 100.0 & 96.87 & 92.91 & 93.77 & 95.14 \\ 
  {\textsc{Comet}}   & 99.3 & 95.77 & 91.04 & 92.18 & 93.74 \\
  + aug   & 98.6 & 95.54 & 91.66 & 92.06 & 93.65 \\
  + SL-feat. & 99.3 & \textbf{96.95} & 93.56 & 94.64 & 95.59 \\
  + WL-tags  & 99.2 & 96.48 & \textbf{93.9} & \textbf{96.36} & \textbf{96.2}\\
\bottomrule
\end{tabular}
}
\addtolength{\tabcolsep}{-0.5pt}
\caption{Accuracy on DEMETR perturbations for both lexical-based and neural-based metrics, shown bucketed by error severity (base, critical, major, and minor errors), including a micro-average across all perturbations.}
\label{tab:demetr2}
\end{table}

One of the key findings from Table~\ref{tab:demetr2} is that the model which uses word-level information consistently outperforms the other methods across almost all severity buckets, with the exception of ``critical'' error bucket. In combination with the findings on the ACES challenge set (see section \ref{sec:aces}), it seems that investigating approaches which target more nuanced and complex error phenomena that lead to critical errors could further improve performance of neural metrics.

\subsubsection{ACES}
\label{sec:aces}
To analyse general, high-level, performance trends of the lexical and proposed approaches on the ACES challenge set, we report Kendall's tau correlation and the ``ACES - Score'' as proposed by \cite{amrhein-etal-2022-aces}, which is a weighted combination of the 10 top-level categories in the ACES ontology:

\begin{equation}
    \text{ACES-Score} = \text{\textit{sum}} \left\{ 
        \begin{array}{ll}
            5 * \tau_{\text{addition}} \\
            5 * \tau_{\text{omission}} \\
            5 * \tau_{\text{mistranslation}} \\
            5 * \tau_{\text{overtranslation}} \\
            5 * \tau_{\text{undertranslation}} \\
            1 * \tau_{\text{untranslated}} \\
            1 * \tau_{\text{do not translate}} \\
            1 * \tau_{\text{real-world knowledge}} \\
            1 * \tau_{\text{wrong language}} \\
            0.1 * \tau_{\text{punctuation}} \\
        \end{array}
    \right\}
\end{equation}

\begin{table*}[t]
\small
\centering
\begin{tabular}{lccccccc}
\toprule
& \multicolumn{1}{c}{\textsc{BLEU}} & \multicolumn{1}{c}{\textsc{chrF}} & \multicolumn{1}{c}{{\textsc{Comet}}} & \multicolumn{1}{c}{\textsc{Ensemble}} & \multicolumn{1}{c}{\textsc{Comet}+aug} & \multicolumn{1}{c}{\textsc{Comet}+SL-feat.} & \multicolumn{1}{c}{\textsc{Comet}+WL-tags} \\
\cmidrule{2-8}  
\multicolumn{8}{c}{\textit{major (weight = 5)}} \\ [0.1cm]
  addition  & \textbf{0.748} & 0.644 & 0.349 & 0.367 & 0.52 & 0.443 & 0.427 \\
  omission  & 0.427 & 0.784 & 0.704 & \textbf{0.828} & 0.706 & 0.724 & 0.666 \\
  mistranslation  & -0.296 & 0.027 & 0.186 & 0.216 & \textbf{0.255} & 0.148 & 0.189 \\ 
  overtranslation  & -0.838 & -0.696 & 0.27 & 0.176 & \textbf{0.308} & 0.086 & 0.304 \\
  undertranslation  & -0.856 & -0.592 & 0.08 & -0.044 & \textbf{0.2} & -0.18 & 0.12 \\
  \hdashline
  \multicolumn{8}{c}{\textit{minor (weight = 1)}} \\ [0.1cm]
  untranslated  & 0.786 & \textbf{0.928} & 0.709 & 0.894 & 0.58 & 0.618   & 0.686 \\
  do not translate  & 0.58 & \textbf{0.96} & 0.88 & 0.9 & 0.78 & 0.9 & 0.84 \\ 
  real-world knowl.  & -0.906 & -0.307 & 0.195 & 0.176 & \textbf{0.202} & 0.109 & 0.162 \\
  wrong language  & 0.659 & \textbf{0.693} & 0.071 & 0.052 & 0.159 & 0.185 & 0.087 \\
  \hdashline 
  \multicolumn{8}{c}{\textit{fluency/punctuation  (weight = 0.1)}} \\ [0.1cm]
  punctuation  & 0.658 & \textbf{0.803} & 0.328 & 0.699 & 0.377 & 0.323 & 0.339 \\
  \midrule 
  ACES-Score  & -2.89 & 3.189 & 9.833 & 9.807 & \textbf{11.704} & 7.949 & 10.339 \\
\bottomrule
\end{tabular}
\addtolength{\tabcolsep}{-0.5pt}
\caption{Kendall’s tau-like correlations for 10 top-level categories in ACES challenge set.}
\label{tab:aces2}
\end{table*}

\begin{table*}[t]
\small
\centering
\begin{tabular}{lccccccc}
\toprule
& \multicolumn{1}{c}{\textsc{BLEU}} & \multicolumn{1}{c}{\textsc{chrF}} & \multicolumn{1}{c}{{\textsc{Comet}}} & \multicolumn{1}{c}{\textsc{Ensemble}} & \multicolumn{1}{c}{\textsc{Comet}+aug} & \multicolumn{1}{c}{\textsc{Comet}+SL-feat.} & \multicolumn{1}{c}{\textsc{Comet}+WL-tags}\\
\cmidrule{2-8}  
  \textsc{En-Xx}  & 0.034 & 0.329 & 0.201 & \textbf{0.340} & 0.256 & 0.183 & 0.206 \\
  \textsc{Xx-En}  & -0.37 & -0.046 & 0.283 & 0.26 & \textbf{0.329} & 0.222 & 0.285 \\
  \textsc{Xx-Yy} & -0.124 & 0.097 & 0.105 & 0.115 & \textbf{0.204} & 0.088 & 0.104 \\
  \midrule 
  AVG   & -0.153 & 0.127 & 0.196 & 0.238 & \textbf{0.263}$^\dagger$ & 0.164 & 0.198 \\
\bottomrule
\end{tabular}
\addtolength{\tabcolsep}{-0.5pt}
\caption{Kendall’s tau-like correlation on ACES challenge set. $\dagger$ in the averaged scores indicates statistically significant difference to the other metrics.}
\label{tab:aces}
\end{table*}

The weights in this formula correspond to the recommended values in the MQM framework \cite{freitag2021experts}: $weight=5$ for major, $weight=1$ for minor and $weight=0.1$ for fluency/punctuation errors. The ACES-Score results can be seen in the last row of Table~\ref{tab:aces2}. 

Overall, as the ACES challenge set contains a larger set of translation errors, and goes beyond simple perturbations to more nuanced error categories such as real-world knowledge and discourse-level errors, we can see that the performance scores and best metrics vary largely depending on the category. Interestingly, \chrf seems to outperform other metrics especially in the categories that do not relate so much to replacements in the reference translation, but rather relate to fully or partially wrong language (or punctuation) use. We note that these seem to be largely cases that are not frequently found in MT training data, nor are they considered in previously proposed data augmentation approaches, which could explain why neural metrics are outperformed by baseline surface-level metrics, even under investigated robustness modifications. Hence, there seems to be room for further improvements in incorporating surface-based information in neural metrics and enabling them to pay more attention to n-gram overlap.
Instead, for the error categories that depend on other perturbations, we can see that all robustness oriented modifications to \comet improve the performance compared to the vanilla model, with augmentation achieving significantly higher Kendall's tau correlations. 

When looking at the overall picture and focusing on the ACES-Score which weights the errors by the severity of the errors there seem to be only two methods that outperform the baseline \textbf{\textsc{Comet}} model, namely \textbf{\textsc{Comet} + aug} and \textbf{\textsc{Comet} + WL-tags}, which achieve the best and second best ACES-Score respectively. Since these two approaches are orthogonal to each other, it seems that a promising direction for future work is to explore options for combining the two methods.

Note that the overall behavior of lexical and neural-based metrics corroborates  the findings presented in the original paper. We can confirm that in our experiments the worst performing metric is also {\textsc{BLEU}}, which is expected. 
However, it is hard to highlight the best performing metric based only on the ACES-Score, the purpose of this analysis is more so to find any interesting trends or any particular issues that some methods are handling better than the others.

Since the ACES dataset encompasses a high number of LPs, we aggregate the results into three groups, \textsc{En-Xx} (out-of-English), \textsc{Xx-En} (into-English) and \textsc{Xx-Yy} (LPs without English). We also report the balanced average across all language pairs (\textsc{Avg}). Results in Table~\ref{tab:aces} show that methods which include augmented data during training achieve higher performance compared to other proposed options. As for additional sentence-level or word-level information, \textbf{\textsc{Comet} + WL-tags} slightly improves performance of the baseline \textsc{Comet} across \textsc{En-Xx} and \textsc{Xx-En} aggregations and beats the approach that uses SL-features.

\section{Conclusion and Future Work}

In this paper, we presented several approaches that use interpretable string-based metrics to improve the robustness of recent neural-based metrics such as \textsc{Comet}. There are various ways of combining these methods together: ensembling metrics, incorporating sentence-level features, or using word-level information coming from alignments between the hypothesis and the reference.  
We observed that adding  small changes to the architecture of {\sc Comet}, either by using sentence-level features based on \textsc{BLEU} and \textsc{chrF} scores,  or by incorporating word-level tags for the hypothesis, can lead to competitive performance gains. 
To showcase the effectiveness of our proposed approaches, we evaluated them on the most recent MQM test set that covers multiple domains and language pairs, as well as on the challenge sets that were introduced in the WMT 2022 Metrics shared task, with encouraging results.

It is likely that our proposed approaches are complementary to each other, as well as to the data augmentation method we are comparing against (COMET+aug). An interesting direction for future work is to  study further the impact of using word-level tags of the hypothesis in other ways not covered in this paper, e.g., in combination with augmentation approaches.

\section*{Acknowledgements}
This work was supported by the European Research Council (ERC StG DeepSPIN 758969), by EU's Horizon Europe Research and Innovation Actions (UTTER, contract 101070631), by P2020 project MAIA (LISBOA-01-0247- FEDER045909), by the Portuguese Recovery and Resilience Plan  through project C645008882-00000055 (NextGenAI, Center for Responsible AI) and Fundação para a Ciência e Tecnologia through contract UIDB/50008/2020.

\bibliography{eamt23}
\bibliographystyle{eamt23}

\clearpage
\newpage
\appendix
\section{Model Implementation and Parameters}
\label{sec:app_model_params}

Table \ref{tab:hp} shows the hyperparameters used to train the following prediction models: \textbf{\textsc{Comet}}, \textbf{\textsc{Comet} + SL-feat.} and \textbf{\textsc{Comet} + WL-tags}. For the baseline we used the code available at \url{https://github.com/Unbabel/COMET} and we trained the model on WMT17-WMT20 DA data (in the table we refer to it as \textbf{\textsc{Comet}}).

For the \textsc{Ensemble} we tune three weights on the development set with grid search, by optimizing Kendall tau correlations (see Table~\ref{tab:weights}). 

\begin{table}[ht!]
\small
\centering
\addtolength{\tabcolsep}{-0.5pt}
\begin{tabular}{lccc}
\toprule
& \textsc{BLEU} & \textsc{chrF} & \textsc{Comet}\\  
\midrule
weights  & 0.02513 & 0.04523 & 0.92965 \\
\bottomrule
\end{tabular}
\addtolength{\tabcolsep}{-0.5pt}
\caption{Tuned weights on the MQM 2021 set for the weighted ensemble.}
\label{tab:weights}
\end{table}


The bottleneck size parameter for \textbf{\textsc{Comet} + SL-feat.} model was tuned using development set. This set covers three language pairs (English $\rightarrow$ German, English $\rightarrow$ Russian, Chinese $\rightarrow$ English) and two domains (ted and newstest). Kendall tau correlation was computed over the whole dataset without considering different domains separately (see Table~\ref{tab:bottleneck_sizes}).

\begin{table}[ht!]
\small
\centering
\begin{tabular}{lcccc}
\toprule
& \multicolumn{1}{c}{\textsc{64}} & \multicolumn{1}{c}{\textsc{128}} & \multicolumn{1}{c}{{\textsc{256}}} & \multicolumn{1}{c}{\textsc{512}} \\
\cmidrule{2-5}  
  \textsc{En-De}  & 0.223 & 0.216 & 0.217 & \textbf{0.225} \\ 
  \textsc{En-Ru}  & \textbf{0.305} & 0.279 & 0.275 & 0.281  \\ 
  \textsc{Zh-En}  & 0.319 & \textbf{0.330} & 0.325 & 0.315 \\ 
  \midrule
  \textsc{AVG}  & \textbf{0.282} & 0.275 & 0.272 & 0.274 \\ 
\bottomrule
\end{tabular}
\addtolength{\tabcolsep}{-0.5pt}
\caption{Kendall’s tau-like correlation per language pair on development set for different bottleneck sizes. \textbf{Bold} values indicate the best performance per language pair.}
\label{tab:bottleneck_sizes}
\end{table}

\begin{table*}[b!]
\small
\centering
\begin{tabular}{lccc}
\toprule 
\textbf{Hyperparameter}  & \textbf{COMET}  & \textbf{COMET + SL-feat.} & \textbf{COMET + WL-tags}   \tabularnewline
\midrule 
Encoder Model  & XLM-R (large)  & XLM-R (large) & XLM-R (large) \tabularnewline
Optimizer  & Adam  & Adam & Adam  \tabularnewline
No. frozen epochs  & 0.3  & 0.3 & 0.3  \tabularnewline
Learning rate  & 3e-05  & 3e-05 & 3e-05 \tabularnewline
Encoder Learning Rate  & 1e-05  & 1e-05 & 1e-05  \tabularnewline  
Layerwise Decay  & 0.95  & 0.95 & 0.95 \tabularnewline
Batch size  & 4  & 4 & 4  \tabularnewline 
Loss function  & Mean squared error  & Mean squared error & Mean squared error \tabularnewline
Dropout  & 0.15  & 0.15 & 0.15 \tabularnewline
Hidden sizes  & [3072, 1024]  & [3072, 1024] & [3072, 1024] \tabularnewline
Encoder Embedding layer & Frozen & Frozen & Frozen  \tabularnewline
Bottleneck layer size & - & 64 & - \tabularnewline
FP precision  & 32  & 32 & 32 \tabularnewline
No. Epochs (training) & 2 & 2 & 2 \\
\bottomrule
\end{tabular}
\caption{Hyperparameters used to train different prediction methods.} 
\label{tab:hp}
\end{table*}

\section{Correlation with Human Judgements}
\label{sec:app_mqm}

We present here results on MQM 2022 set for Pearson and spearman correlations (see Tables~\ref{tab:mqm_pearson} and ~\ref{tab:mqm_spearman} accordingly). We can see that especially for Spearman $\rho$ the findings are aligned with the findings on Kendall tau correlations. Instead, for the Pearson $r$ which is more sensitive to outliers, we can see that the augmentation method outperforms the feature-based modifications.

\begin{table*}[ht!]
\small
\centering
\begin{tabular}{clccccccc}
\toprule
& & \multicolumn{1}{c}{\textsc{Bleu}} & \multicolumn{1}{c}{\textsc{chrf}} & \multicolumn{1}{c}{\textsc{Comet}} & \multicolumn{1}{c}{\textsc{ensemble}} & \multicolumn{1}{c}{\textsc{Comet} + aug} & \multicolumn{1}{c}{+ SL-feat.} & \multicolumn{1}{c}{ + WL-tags} \\
\cmidrule{3-9}  

\multirow{5}{*}{\rotatebox{90}{\textsc{En-De}}} 
  & Conversation  & 0.228 & 0.285 & 0.371 & 0.376 & 0.378 & 0.379 & \textbf{0.400} \\
  & Ecommerce  & 0.173 & 0.222 & 0.376 & 0.373 & 0.380 & \textbf{0.383} & 0.341 \\
  & News & 0.220 & 0.260 & 0.521 & 0.521 & 0.492 & 0.506 & \textbf{0.526} \\
  & Social & 0.172 & 0.220 & 0.367 & 0.367 & 0.375 & \textbf{0.382} & 0.351 \\
  \midrule
\multirow{5}{*}{\rotatebox{90}{\textsc{En-Ru}}}  
  & Conversation  & 0.155 & 0.185 & 0.372 & 0.369 & \textbf{0.418} & 0.350 & 0.400 \\
  & Ecommerce  & 0.249 & 0.287 & 0.488 & 0.488 & \textbf{0.510} & 0.507 & 0.481 \\
  & News & 0.169 & 0.230 & \textbf{0.469} & 0.467 & 0.464 & 0.477 & 0.448 \\
  & Social & 0.213 & 0.143 & 0.324 & 0.328 & 0.371 & 0.343 & \textbf{0.385} \\
  \midrule
  \multirow{5}{*}{\rotatebox{90}{\textsc{Zh-En}}}  
  & Conversation  & 0.160 & 0.206 & 0.340 & 0.338 & \textbf{0.370} & 0.343 & 0.358 \\
  & Ecommerce  & 0.220 & 0.230 & 0.391 & 0.391 & 0.438 & 0.400 & \textbf{0.440} \\
  & News & 0.097 & 0.078 & 0.340 & 0.334 & \textbf{0.383} & 0.364 & 0.359 \\
  & Social & 0.161 & 0.177 & 0.351 & 0.347 & 0.358 & 0.343 & \textbf{0.373} \\
  \midrule 
\multirow{5}{*}{\rotatebox{90}{\textsc{ }}}  
  & AVG  & 0.185 & 0.210 & 0.393 & 0.392 & \textbf{0.411} & 0.398 & 0.405 \\
\bottomrule
\end{tabular}
\addtolength{\tabcolsep}{-0.5pt}
\caption{Pearson correlation on high resource language pairs using the MQM annotations for Conversation, Ecommerce, News and Social domains collected for the WMT 2022 Metrics Task. \textbf{Bold} numbers indicate the best result for each domain in each language pair.}
\label{tab:mqm_pearson}
\end{table*}

\begin{table*}[ht!]
\small
\centering
\begin{tabular}{clcccccccc}
\toprule
& & \multicolumn{1}{c}{\textsc{Bleu}} & \multicolumn{1}{c}{\textsc{chrf}} & \multicolumn{1}{c}{\textsc{Comet}} & \multicolumn{1}{c}{\textsc{ensemble}} & \multicolumn{1}{c}{\textsc{Comet} + aug} & \multicolumn{1}{c}{+ SL-feat.} & \multicolumn{1}{c}{ + WL-tags}\\
\cmidrule{3-10}  
\multirow{5}{*}{\rotatebox{90}{\textsc{En-De}}} 
  & Conversation  & 0.262 & 0.337 & 0.401 & 0.403 & 0.385 & 0.404 & \textbf{0.409}  \\
  & Ecommerce  & 0.235 & 0.278 & \textbf{0.421} & 0.411 & 0.403 & 0.416 & 0.417  \\
  & News & 0.224 & 0.273 & 0.478 & 0.472 & 0.438 & 0.471 & \textbf{0.486}  \\
  & Social & 0.173 & 0.222 & \textbf{0.389} & 0.383 & 0.361 & 0.386 &  0.384 \\
  \midrule
\multirow{5}{*}{\rotatebox{90}{\textsc{En-Ru}}}  
  & Conversation  & 0.183 & 0.230 & 0.400 & 0.397 & 0.427 & 0.389 &  \textbf{0.428} \\
  & Ecommerce  & 0.276 & 0.303 & 0.502 & 0.501 & 0.514 & 0.499 & \textbf{0.528}  \\
  & News & 0.171 & 0.224 & 0.499 & 0.492 & 0.490 & \textbf{0.514} &  0.495 \\
  & Social & 0.212 & 0.186 & 0.425 & 0.423 & 0.455 & 0.460 &  \textbf{0.483} \\
  \midrule
  \multirow{5}{*}{\rotatebox{90}{\textsc{Zh-En}}}  
  & Conversation  & 0.166 & 0.211 & 0.375 & 0.369 & 0.385 & 0.370 & \textbf{0.389} \\
  & Ecommerce  & 0.241 & 0.259 & 0.449 & 0.448 & 0.467 & 0.459 & \textbf{0.487}  \\
  & News & 0.063 & 0.057 & 0.364 & 0.352 & 0.393 & 0.373 &  \textbf{0.394} \\
  & Social & 0.219 & 0.256 & 0.424 & 0.421 & 0.418 & 0.419 & \textbf{0.439}  \\
  \midrule 
\multirow{5}{*}{\rotatebox{90}{\textsc{ }}}  
  & AVG   & 0.202 & 0.236 & 0.427 & 0.423 & 0.428 & 0.430 & \textbf{0.445}  \\
\bottomrule
\end{tabular}
\addtolength{\tabcolsep}{-0.5pt}
\caption{Spearman correlation on high resource language pairs using the MQM annotations for Conversation, Ecommerce, News and Social domains collected for the WMT 2022 Metrics Task. \textbf{Bold} numbers indicate the best result for each domain in each language pair.}
\label{tab:mqm_spearman}
\end{table*}

\section{Training Data Statistics}
\label{sec:data_stats}

The combined WMT training data (from 2017 to 2020) has 950069 segments and covers the following language pairs (total number is 32): Cs-En, De-Cs, De-En, De-Fr, En-Cs, En-De, En-Et, En-Fi, En-Gu, En-Ja, En-Kk, En-Lt, En-Lv, En-Pl, En-Ru, En-Ta, En-Tr, En-Zh, Et-En, Fi-En, Fr-De, Gu-En, Ja-En, Kk-En, Km-En, Lt-En, Pl-En, Ps-En, Ru-En, Ta-En, Tr-En, Zh-En.

\end{document}